\documentclass[final,twocolumn,9pt]{IEEEtran}

\usepackage{spconf,graphicx}

\usepackage{amsmath,epsfig}

\usepackage{comment}
\usepackage{url} 
\usepackage{hyperref}
\usepackage{graphics}

\usepackage{multirow}

\title{I-HAZE: a dehazing benchmark with real hazy and haze-free indoor images}

\name{Codruta O. Ancuti $^{*}$, Cosmin Ancuti $^{*}$,  Radu Timofte $^{\ddag}$ and Christophe De Vleeschouwer $^{\dag}$ }
\address{ $^{*}$  MEO, Universitatea Politehnica Timisoara, Romania\\
$^{\ddag}$ ETH Zurich, Switzerland and Merantix GmbH, Germany\\
$^{\dag}$ ICTEAM, Universite Catholique de Louvain, Belgium\\
}

\begin{document}
%
\maketitle

\begin{abstract}
Image dehazing has become an important computational imaging topic in the recent years. However, due to the lack of ground truth images, the comparison of dehazing methods is not straightforward, nor objective. To overcome this issue we introduce I-HAZE, a new dataset that contains 35 image pairs of hazy and corresponding haze-free (ground-truth) indoor images. Different from most of the existing dehazing databases, hazy images have been generated using real haze produced by a professional haze machine. To ease color calibration and improve the assessment of dehazing algorithms, each scene include a MacBeth color checker. Moreover, since the images are captured in a controlled environment, both haze-free and hazy images are captured under the same illumination conditions. This represents an important advantage of the I-HAZE dataset that allows us to objectively compare the existing image dehazing techniques using traditional image quality metrics such as PSNR and SSIM. 
\end{abstract}
%


\section{Introduction}

Limited visibility and reduced contrast due to haze or fog conditions is a major issue that hinders the success of many outdoor computer vision and image processing algorithms. Consequently, automatic dehazing methods have been largely investigated.  Oldest approaches rely on atmospheric cues~\cite{Cozman_Krotkov_97,Narasimhan_2002}, multiple images captured with polarization filters~\cite{PAMI_2003_Narasimhan_Nayar,Schechner_2003}, or known depth~\cite{Kopf_DeepPhoto_SggAsia2008,Tarel_ICCV_2009}. Single image dehazing, meaning dehazing without side information related to the scene geometry or to the atmospheric conditions, is a complex mathematically ill-posed problem. This is because the degradation caused by haze is different for every pixel
and depends on the distance between the scene point and the camera. This dependency is expressed in the transmission coefficients, that control the scene attenuation and amount of haze in every pixel. Due to lack of space, we refer the reader to previous papers for a formal description of a simplified but realistic light propagation model, combining transmission and airlight to describe how haze impacts the observed image.

\begin{figure}[t]
  \centering
  \includegraphics[width=1\linewidth]{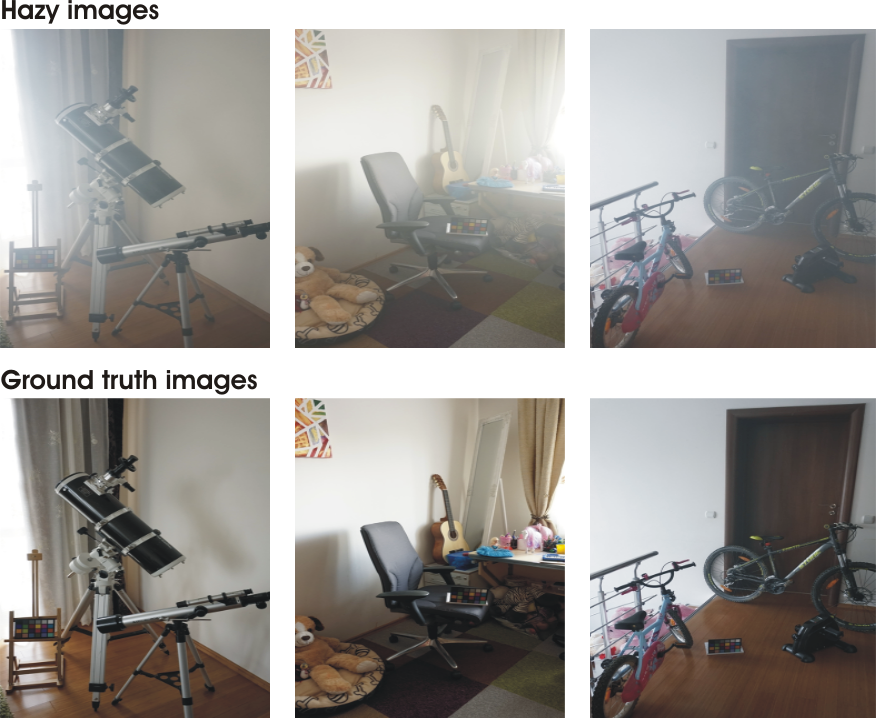}
  \caption{\label{fig:intro}%
     \textit{I-HAZE dataset provides 35 set of hazy  indoor images and the corresponding ground truth (haze-free) images.} 
  }  
\end{figure}

Single image dehazing directly builds on this simplified model. It has been addressed recently~\cite{Fattal_Dehazing,Tan_Dehazing,Dehaze_He_CVPR_2009,Tarel_ICCV_2009,Kratz_and_Nishino_2009,Dehaze_Ancuti_ACCV,Codruta_ICIP_2010,Ancuti_TIP_2013,Fattal_Dehazing_TOG2014,Ancuti_GRSL_2014,Emberton_2015,Tang_2014}, by considering different kinds of priors to estimate the transmission. 
The method of Fattal~\cite{Fattal_Dehazing} adopts a refined image formation model that accounts for surface shading in addition to the transmission function. This allows to regularize the transmission and haze color estimation problems by searching for a solution in which the resulting shading and transmission functions are locally statistically uncorrelated. Tan's approach~\cite{Tan_Dehazing} assumes the atmospheric airlight to be the brightest pixel in the scene, and estimate the transmission by maximizing the contrast. The maximal contrast assumption has also been exploited in~\cite{Tarel_ICCV_2009}, with a linear complexity in the number of image pixels. The dark channel priors, introduced in~\cite{Dehaze_He_CVPR_2009},  has been proven to be a really effective solution to estimate the transmission, and has motivated many recent approaches. Meng et al.~\cite{Meng_2013} extends the work on dark channel, by regularizing the transmission around boundaries to mitigate its lack of resolution. Zhu et al.~\cite{Zhu_2015} extend~\cite{Meng_2013} by considering a color attenuation prior, assuming that the depth can be estimated from pixel saturation and intensity. The color-lines, introduced in~\cite{Fattal_Dehazing_TOG2014}, also exploit the impact of haze over the color channels distribution. Berman et al.~\cite{Berman_2016} adopt a similar path. They observe that the colors of a haze-free image are well approximated by a limited set of tight clusters in the RGB space. In presence of haze, those clusters are spread along lines in the RGB space, as a function of the distance map, which allows to recover the haze free image. Ancuti et al.~\cite{Dehaze_Ancuti_ACCV} rely on the hue channel analysis to identify hazy regions and enhance the image. For night-time dehazing and non-uniform lighting conditions, spatially varying airlight has been considered and estimated in~\cite{Ancuti_NightTime, Li_2015}. 
Fusion-based single image dehazing approaches~\cite{Ancuti_TIP_2013,Choi_2015} have also achieved visually pleasant results without explicit transmission estimation and, more recently, several machine learning based methods have been introduced~\cite{Tang_2014,Dehazenet_2016,Ren_2016}. DehazeNet~\cite{Dehazenet_2016} takes a hazy image as input, and outputs its medium transmission map that is subsequently used to recover a haze-free image via atmospheric scattering model. It resort  to  synthesized training data based on the physical haze formation model. Ren et al.~\cite{Ren_2016} proposed a coarse-to-fine network consisting of a cascade of CNN layers, also trained with synthesized hazy images.

Despite this prolific set of dehazing algorithms, the validation and comparison of those methods remains largely unsatisfactory.
Due to the absence of corresponding pairs of hazy and haze-free ground-truth image, most of the existing evaluation methods are based on non-reference image quality assessment (NR-IQA) strategies. For example, in~\cite{Hautiere_2008}, the assessment simply relies on the gradient of the visible edges. Other non-reference image quality assessment (NR-IQA) strategies~\cite{Mittal_2012,Mittal_2013,Saad_2012} have been used for dehazing assessment. A more general framework has been introduced in~\cite{Chen_2014}, using subjective assessment of enhanced and original images captured in bad visibility conditions. Besides, the Fog Aware Density Evaluator (FADE) introduced in~\cite{Choi_2015} predicts the visibility of a hazy/foggy scene from a single image without corresponding ground-truth. 
Unfortunately, due to the absence of the references (haze-free), none of these quality assessment approaches has been commonly accepted by the dehazing community.

Due to the practical issues associated to the recording of reference and hazy images under identical illumination condition, all existing data-sets have been built on synthesized hazy images based on the optical model and known depth.  
The work~\cite{Tarel_2012} presents the FRIDA dataset designed for Advanced Driver Assistance Systems (ADAS) that is a synthetic image database (computer graphics generated scenes) with 66 roads synthesized scenes. In~\cite{D_Hazy_2016}, a dataset of 1400+ images of real complex scenes has been derived from the \textit{Middleburry}\footnote{\scriptsize{\url{http://vision.middlebury.edu/stereo/data/scenes2014/}}}  and the \textit{NYU-Depth V2}\footnote{\scriptsize{\url{http://cs.nyu.edu/~silberman/datasets/nyu_depth_v2.html}}} datasets. It contains high quality real scenes, and the depth map associated to each image has been used to yield synthesized hazy images based on Koschmieder's light propagation model~\cite{Koschmieder_1924}. This dataset has been recently extended in~\cite{HazeRD_2017}.

Complementary to the existing dehazing datasets, in this paper we contribute I-HAZE, a new dataset containing pairs of real hazy and corresponding haze-free images for $35$ various indoor scenes.
Haze has been generated with a professional haze machine that imitates with high fidelity real hazy conditions. A similar dataset (CHIC) that contains only two quite similar scenes was introduce in~\cite{Khoury_2016}. However, I-HAZE dataset provides significantly more scenes with a larger variety of object structures and colors. 
Another contribution of this paper is a comprehensive evaluation of several state-of-the-art single image dehazing methods. 
Interestingly, our work reveals that many of the existing dehazing techniques are not able to accurately reconstruct the original image from its hazy version. This observation is founded on SSIM~\cite{Wang_2006} and CIEDE2000~\cite{Sharma_2005} image quality metrics, computed using the known reference and the dehazed results produced by different dehazing techniques. This observation, combined with the release of our dataset, certainly paves the way for improved dehazing methods.


\section{I-HAZE Dataset}

This section describes how the I-Haze dataset has been produced.
All the 35 scenes presented in the I-Haze correspond to indoor domestic environments, with objects with different colors and specularities. Besides the domestic objects,  all the scenes contains a color checker chart (Macbeth color checker). We use a classical Macbeth color checker with the size 11 by 8.25 inches with 24 squares of painted samples (4x6 grid).

After carefully setting each scene, we first record the ground truth (haze-free image) and then we immediately start the process of introducing haze in the scene. Therefore, we use two professional fog/haze machines (LSM1500 PRO 1500 W) that generate a dense vapor.  The fog generators use cast or platen type aluminum heat exchangers, which causes evaporation of the water-based fog liquid.  The generated  particles (since are water droplets) have approximately the same diameter size of 1 - 10 microns as the atmospheric haze. Before shooting the hazy scene, we use a fan that helps to obtain in a relatively short period of time a homogenous  haze distribution in the entire room (obviously  that is kept isolated as much as possible by closing all the doors and windows). The entire process to generate haze took approximately 1 minute. Waiting approximately another 5-10 minutes, we obtain a homogenous distribution of the haze. The distances between the camera and the target objects range form 3 to 10 meters. The recordings were performed during the daytime in relatively short intervals (20-30 minutes per scene recording) with natural lightning and when the light remains relatively constant (either smooth cloudy days or when the sun beams did not hit directly the room windows). 

To capture haze-free and hazy images, we used a setup that includes a tripod and a Sony A5000 camera that was remotely controlled (Sony RM-VPR1). We acquired JPG and ARW (RAW) 5456$\times$3632 images, with 24 bit depth. 
The cameras were set on manual mode and we kept the camera still (on a tripod) over the entire shooting session of the scene.  We calibrate the camera in haze-free scene, and then we kept the same parameters for the hazy scene. For each scene, the camera settings has been manually calibrated by adjusting manually the aperture (F-stop), shutter-speed (exposure-time), ISO speed and the white-balance.  Setting the three parameters aperture-exposure-ISO has been realized using both the built-in light-meter of the camera and an external exponometer Sekonic.  For the white-balance we used the gray-card, targeting a middle gray (18$\%$ gray). The calibration process is straight-forward, since it just requires to set the white-balance in manual mode and to place the gray-card in front of the subject. In practice, we placed the gray-card in the center of the scene, two meters away from the camera.  In addition, since each scene contains the color checker, the white-balance can be manually adjusted (a posteriori) using specialized software such as Adobe Photoshop Lightroom.


\section{Quantitative Evaluation}

\begin{figure*}[t!]
  \centering
  \includegraphics[width=1\linewidth]{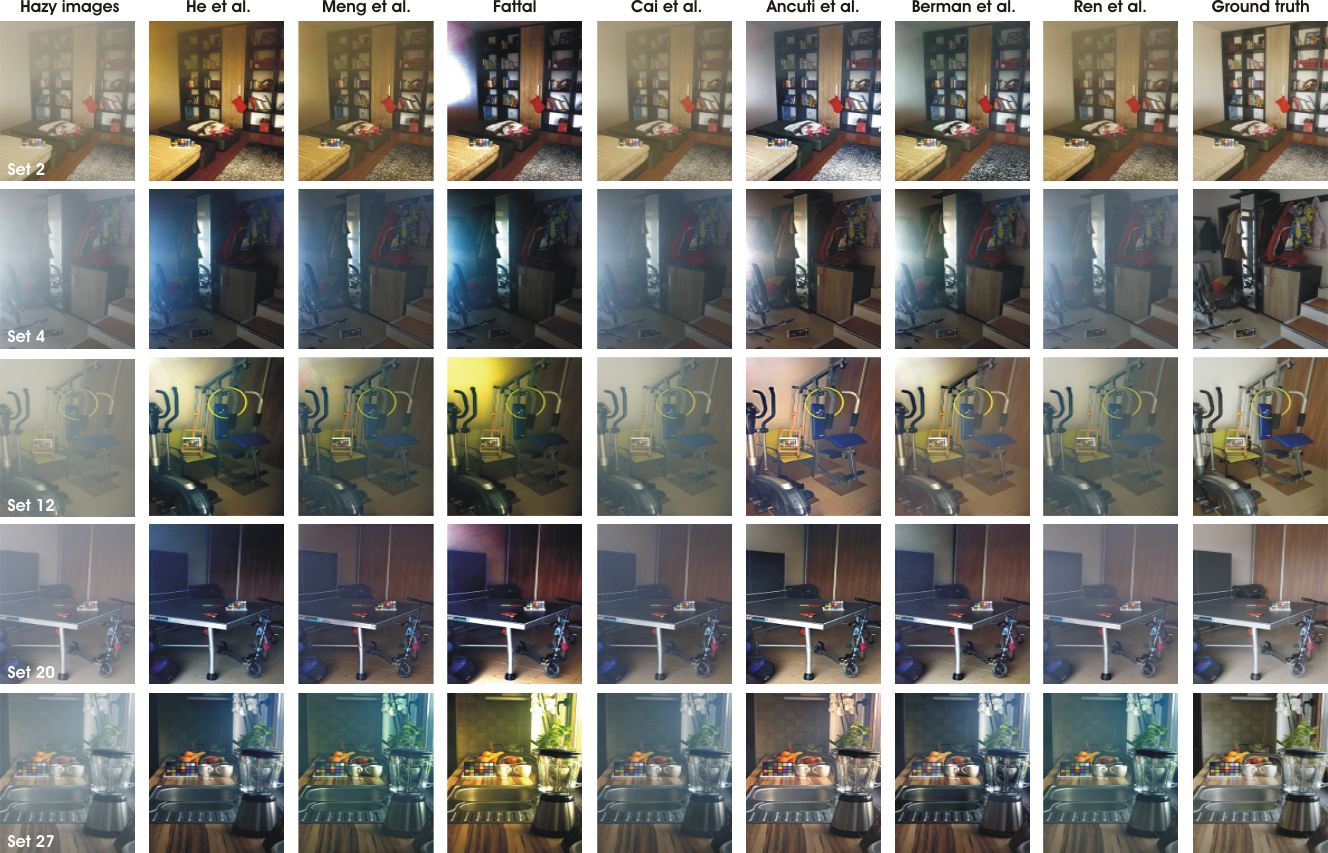}
  \caption{\label{fig:res_comp1}%
    \textit{\textbf{Comparative results.} The first row shows the hazy images and the last row shows  the ground truth. The other rows from left to right show the results of He et al.~\cite{Dehaze_He_CVPR_2009}, Meng et al.~\cite{Meng_2013}, Fattal~\cite{Fattal_Dehazing_TOG2014}, Cai et al.~\cite{Dehazenet_2016}, Ancuti et al.~\cite{Ancuti_NightTime},  Berman et al.~\cite{Berman_2016} and Ren et al.~\cite{Ren_2016}.}
  }  
\end{figure*}

In this section, the I-HAZE dataset is used to perform a comprehensive quantitative evaluation of several state-of-the-art single image based dehazing techniques. Before presenting the techniques used in our evaluation we briefly discuss the optical model assumed in most of the existing dehazing techniques.

Mathematically the dehazing optical model is expressed by the image formation model of Koschmieder~\cite{Koschmieder_1924}. Based on this model, due to the atmospheric particles that absorb and scatter light, only a certain percentage of the reflected light reaches the observer. The light intensity $\mathcal{I}$  of each pixel coordinate $x$, that passes a hazy medium is expressed as:
\begin{equation}
	\mathcal{I}(x)=\mathcal{J}(x) \enspace T \left(x \right)+A_\infty \enspace \left[ 1-T\left(x\right) \right]
\label{model}
\end{equation} 
where the haze-free image is denoted by $\mathcal{J}$, $T$ is the \textit{transmission} (depth map) and  $A_\infty$ is the atmospheric light (a color constant).\newline

\begin{table*}[t!]
\centering
\label{table_1}
\begin{tabular}{|l|r|r|r|r|r|r|r|r|r|r|r|r|r|r|}
\hline
\multirow{2}{*}{} & \multicolumn{2}{c|}{\textbf{He et al.}\cite{Dehaze_He_CVPR_2009}} & \multicolumn{2}{c|}{\textbf{Meng et al.}\cite{Meng_2013}} & \multicolumn{2}{c|}{\textbf{Fattal}\cite{Fattal_Dehazing_TOG2014}} & \multicolumn{2}{c|}{\textbf{Cai et al.}\cite{Dehazenet_2016}} & \multicolumn{2}{c|}{\textbf{Ancuti et al.}\cite{Ancuti_NightTime}} & \multicolumn{2}{c|}{\textbf{Berman et al.}\cite{Berman_2016}} & \multicolumn{2}{c|}{\textbf{Ren et al.}\cite{Ren_2016}} \\ \cline{2-15}
                  & {\scriptsize{SSIM}}             & {\tiny{CIEDE2000}}            & {\scriptsize{SSIM}}              & {\tiny{CIEDE2000}}            & {\scriptsize{SSIM}}            & {\tiny{CIEDE2000}}          & {\scriptsize{SSIM}}             & {\tiny{CIEDE2000}}            & {\scriptsize{SSIM}}               & {\tiny{CIEDE2000}}             & {\scriptsize{SSIM}}               & {\tiny{CIEDE2000}}              & {\scriptsize{SSIM}}              & {\tiny{CIEDE2000}}            \\ \hline
\textbf{Set 2}    & 0.783            & 15.665               & 0.815             & 13.577               & 0.607           & 19.849             & 0.849             & 9.806                & 0.824              & 14.073                 & 0.802              & 14.127                 & 0.848             & 9.909                \\ \hline
\textbf{Set 4}    & 0.633            & 20.767               & 0.700             & 16.579               & 0.568           & 20.920             & 0.588             & 19.969               & 0.742              & 14.472                 & 0.691              & 17.597                 & 0.643             & 18.063               \\ \hline
\textbf{Set 12}   & 0.735            & 15.788               & 0.781             & 16.392               & 0.703           & 18.972             & 0.758             & 17.322               & 0.844              & 13.229                 & 0.827              & 9.574                  & 0.829             & 11.810               \\ \hline
\textbf{Set 20}   & 0.617            & 24.836               & 0.790             & 19.568               & 0.539           & 23.428             & 0.608             & 24.042               & 0.763              & 15.763                 & 0.806              & 16.010                 & 0.899             & 10.737               \\ \hline
\textbf{Set 27}   & 0.778            & 18.028               & 0.819             & 17.251               & 0.818           & 15.876             & 0.631             & 20.863               & 0.883              & 10.518                 & 0.788              & 16.867                 & 0.867             & 13.780               \\ \hline
\end{tabular}
\caption{\label{tabel_eval1} \textit{\textbf{Quantitative evaluation.} We randomly selected five set of images from our I-HAZE dataset and  we compute the  SSIM and CIEDE2000 indexes between the ground truth images and the enhanced results of the evaluated techniques. The hazy images, ground truth and the results are shown in Fig.\ref{fig:res_comp1}.}}
\end{table*}

\begin{table*}[t!]
\centering
\label{table_average}
\begin{tabular}{lllllllllll}
\cline{1-8}
\multicolumn{1}{|l|}{}                   & \multicolumn{1}{l|}{\textbf{He et al.}~\cite{Dehaze_He_CVPR_2009}} & \multicolumn{1}{l|}{\textbf{Meng et al.}~\cite{Meng_2013}} & \multicolumn{1}{l|}{\textbf{Fattal}~\cite{Fattal_Dehazing_TOG2014}} & \multicolumn{1}{l|}{\textbf{Cai et al.}~\cite{Dehazenet_2016}} & \multicolumn{1}{l|}{\textbf{Ancuti et al.}~\cite{Ancuti_NightTime}} & \multicolumn{1}{l|}{\textbf{Berman et al.}~\cite{Berman_2016}} & \multicolumn{1}{l|}{\textbf{Ren et al.}~\cite{Ren_2016}} &  &  &  \\ \cline{1-8}
\multicolumn{1}{|l|}{\textbf{SSIM}}      & \multicolumn{1}{r|}{0.711}              & \multicolumn{1}{r|}{0.750}               & \multicolumn{1}{r|}{0.574}           & \multicolumn{1}{r|}{0.6397}               & \multicolumn{1}{r|}{0.770}                  & \multicolumn{1}{r|}{0.767}                  & \multicolumn{1}{r|}{0.791}               &  &  &  \\ \cline{1-8}
\multicolumn{1}{|l|}{\textbf{PSNR}}      & \multicolumn{1}{r|}{15.285}             & \multicolumn{1}{r|}{14.574}              & \multicolumn{1}{r|}{12.421}          & \multicolumn{1}{r|}{14.329}              & \multicolumn{1}{r|}{16.632}                 & \multicolumn{1}{r|}{15.942}                 & \multicolumn{1}{r|}{17.280}              &  &  &  \\ \cline{1-8}
\multicolumn{1}{|l|}{\textbf{CIEDE2000}} & \multicolumn{1}{r|}{17.171}             & \multicolumn{1}{r|}{16.834}              & \multicolumn{1}{r|}{21.385}          & \multicolumn{1}{r|}{17.114}              & \multicolumn{1}{r|}{14.428}                 & \multicolumn{1}{r|}{14.629}                 & \multicolumn{1}{r|}{12.736}              &  &  &  \\ \cline{1-8}
\end{tabular}
\caption{Quantitative evaluation of all the 35 set of images. In this table are shown the average values of the SSIM, PSNR and CIEDE2000 indexes  over the entire dataset. }
\end{table*}

\noindent\textbf{He et al.~\cite{Dehaze_He_CVPR_2009,Dark_Ch_PAMI_2011}} introduced the popular dark-channel prior, an extension of the dark object assumption~\cite{Chavez_1988}. Their proposed dehazing algorithm exploits the observation that the majority of local regions (except the sky, or hazy regions) include some pixels that are characterized by a very low value in at least one color channel. This helps in roughly estimating the transmission map of the hazy images. Further refinement of the transmission map can be obtained based on an alpha matting strategy~\cite{Dehaze_He_CVPR_2009}, or by using guided filters ~\cite{Dark_Ch_PAMI_2011}. In our assessment, we employ the dark channel prior refined based on the guided filter~\cite{Guided_filter_PAMI_2013}. \newline
\textbf{Meng et al.~\cite{Meng_2013}} extend the dark channel prior~\cite{Dehaze_He_CVPR_2009}. It estimates the transmission map by formulating an optimization problem that embeds the constraints imposed by the scene radiance and a weighted L1-norm based contextual regularization, to avoid halo artifacts around sharp edges.
\newline
\textbf{Fattal~\cite{Fattal_Dehazing_TOG2014}} introduced a method  that exploits the observation that pixels of small image patches typically exhibit a one-dimensional distribution in RGB color space, named color lines. Since the haze tends to move color lines away from the RGB origin, an initial estimation of the transmission map is obtained by computing the lines offset from the origin. The final transmission map is refined by applying a Markov random field, which filters the noise and other artifacts resulting from the scattering.
\newline
\textbf{Cai et al.~\cite{Dehazenet_2016}} introduced an end-to-end system built on CNN that learns the mapping relations between hazy and haze free corresponding patches. To train the network they syntheses hazy images based on Middlebury stereo dataset.  The method employs a non  linear activation function that uses a bilateral restraint to  improve convergence. 
\newline
\textbf{Ancuti et al.~\cite{Ancuti_NightTime}} introduced a local airlight estimation applied in a multi-scale fusion strategy for single-image dehazing. The method has been designed to solve the problems associated to the scattering effect, which is especially significant in the nigh-time hazy scenes. The method, however, generalizes to day-time dehazing.  
\newline
\textbf{Berman et al.~\cite{Berman_2016}} introduced an algorithm that also use local airlight estimation, and extends the  color consistency observation of Fattal~\cite{Fattal_Dehazing_TOG2014}.  The algorithm builds on the observation that the colors of a haze-free image can be approximated by a few distinct tight color clusters. Since the pixels of a cluster are spread on the whole image, they are affected differently by haze, to be spread along an elongated line, named haze-line. Those lines convey information about the transmission in different regions of the image, and are thus used for transmission estimation. 
\newline
\textbf{Ren et al.~\cite{Ren_2016}} adopts a multi-scale convolutional neural network to learn the mapping between hazy images and their corresponding transmission maps. The network is trained based on synthetic hazy images, generated by applying a simplified light propagation model to haze-free images for which corresponding depth maps are known. 
\newline


\section{Results and Discussion}

In Fig.~\ref{fig:res_comp1} are shown five hazy images (first column), the corresponding ground truth (last column) and the dehazing results yielded by the specialised techniques of He et al.~\cite{Dehaze_He_CVPR_2009}, Meng et al.~\cite{Meng_2013}, Fattal~\cite{Fattal_Dehazing_TOG2014}, Cai et al.~\cite{Dehazenet_2016}, Ancuti et al.~\cite{Ancuti_NightTime},  Berman et al.~\cite{Berman_2016} and Ren et al.~\cite{Ren_2016}.

Qualitatively, the well-known  method of  He et al.~\cite{Dehaze_He_CVPR_2009} yields results that visually seems to recover the structure, but suffers from color shifting in the hazy regions due to the poor airlight estimation. This happens when the scenes contains lighter color patches in the close-up regions or small reflections. 
The Meng et al.~\cite{Meng_2013} approach, built also on dark channel prior, as expected, generates similar results as He et al.~\cite{Dehaze_He_CVPR_2009}. It presents an improved alternative filtering of the transmission for artifacts reduction and a more precise airlight estimation. 
The results of color-lines method of~\cite{Fattal_Dehazing_TOG2014} suffers from unpleasing color shifting while the results of Ancuti et al.~\cite{Ancuti_NightTime} are more accurate due to the local airlight estimation strategy. This approach also is less prone to introduce structural artifacts due to the  multi-scale fusion strategy.  
The results of Berman et al~\cite{Berman_2016} are yielding visually compelling results and although it is built on a haze-line strategy due to its locally estimation of the airlight,  shown to introduce  less color shifting than~\cite{Fattal_Dehazing_TOG2014}.   
In addition to the algorithms built on priors, deep learning techniques produce less visual artifacts. The results of Ren et al~\cite{Ren_2016} generates visually more compelling results in comparison with the ones generated by DehazeNet of Cai et al~\cite{Dehazenet_2016}. 

To quantitatively evaluate the dehazing methods described in the previous section, we compare directly their outcome with the ground-truth (haze free) images. Besides the well known PSNR,  we compute also the structure similarity index SSIM~\cite{Wang_2004} that compares local patterns of pixel intensities that have been normalized for luminance and contrast. The structure similarity index yields values in the range [-1,1] with maximum value 1 for two identical images. Additionally, to evaluate also the color restoration we employ the CIEDE2000~\cite{Sharma_2005,Westland_2012}.  Different than the earlier measures (e.g. CIE76 and CIE94) that shown shown important limitations to resolve the perceptual uniformity issue, CIEDE2000 defines a more complex, yet most accurate color difference algorithm. CIEDE2000 yields values in the range [0,100] with smaller values indicating better color preservation.

Table~\ref{tabel_eval1}  presents a detailed validation based on SSIM and CIEDE2000 for five randomly selected images of the I-HAZE dataset that are shown in the Fig.~\ref{fig:res_comp1}. In Table~\ref{table_average} are presented the average values over the entire dataset of the SSIM, PSNR and CIEDE indexes.\newline

From  these tables, we can conclude that the methods of Berman et al.~\cite{Berman_2016}, Ancuti et al.~\cite{Ancuti_NightTime}  and Ren et al.~\cite{Ren_2016} performs the best in average when considering the SSIM, PSNR and CIEDE indexes. A second group of methods including Meng et al.~\cite{Meng_2013} and He et al.~\cite{Dehaze_He_CVPR_2009}, perform relatively well both in terms of structure and color restoration.\newline

In general, all the tested methods introduce structural distortions such as halo artifacts close to the edges, that are amplified in the faraway regions. Moreover, due to the poor estimation of the airlight  and transmission map from the hazy image, some color distortions may create some unnatural appearance of the restored images. 
In summary, there is not a single technique that performs the best for all images. The relatively low values of  SSIM, PSNR and CIEDE2000 measures  prove once again the difficulty of single image dehazing task and the fact    there is still much room for improvement.\newline

\bibliographystyle{IEEEbib}
\bibliography{ref_IHAZE}

\end{document}